\documentclass[12pt]{article}
\usepackage{amsmath}
\usepackage{amsfonts}
\usepackage{amssymb}

\usepackage{comment}

\usepackage{graphicx}
\usepackage{enumerate}
\usepackage{natbib}
\usepackage{url} % not crucial - just used below for the URL 

\usepackage{bbm}  % for mathblackboard form of indicator function
\newcommand{\1}{\mathbbm{1}}
\newcommand{\E}{\mathbbm{E}}
\newcommand{\V}{\mathbbm{V}}
\usepackage{bm}
\newcommand{\mb}{\boldsymbol}
\usepackage[pdftex]{hyperref}

\begin{document}

\def\spacingset#1{\renewcommand{\baselinestretch}%
{#1}\small\normalsize} \spacingset{1}

%%%%%%%%%%%%%%%%%%%%%%%%%%%%%%%%%%%%%%%%%%%%%%%%%%%%%%%%%%%%%%%%%%%%%%%%%%%%%%

{
  \title{\bf Hierarchical Gaussian Process Models for Regression Discontinuity/Kink under Sharp and Fuzzy Designs}
  \author{Ximing Wu\thanks{Department of Agricultural Economics, Texas A\&M University, College Station, TX 77843; 
 email: xwu@tamu.edu}}
\date{\empty}
    \maketitle
} 

\thispagestyle{empty}

\bigskip
\begin{abstract}
We propose  nonparametric Bayesian estimators for causal inference exploiting Regression Discontinuity/Kink (RD/RK) under sharp and fuzzy designs. Our estimators are based on  Gaussian Process (GP) regression and classification.  The GP methods are powerful probabilistic machine learning approaches that are advantageous in terms of derivative estimation and uncertainty quantification, facilitating RK estimation and inference of RD/RK models. These estimators are extended to  hierarchical GP models with an intermediate Bayesian neural network layer and can be characterized as  hybrid deep learning models. Monte Carlo simulations show that our estimators perform comparably to and sometimes better than competing estimators in terms of precision, coverage and interval length. The hierarchical GP models considerably improve upon one-layer GP models. We apply the proposed methods to estimate the incumbency advantage of US house elections. Our estimations suggest a significant incumbency advantage in terms of both vote share and probability of winning in the next elections. Lastly we present an extension to accommodate covariate adjustment.
\end{abstract}
\vskip .5in
\noindent%
{\it Keywords:}  Regression Discontinuity/Kink; Sharp and Fuzzy Discontinuity Designs; Nonparametric Bayesian Estimation;  US House Elections
\vfill

\newpage
\spacingset{1.6} % DON'T change the spacing!

\clearpage
\setcounter{page}{1}

\section{Introduction}

The regression discontinuity design \citep{thistlethwaite1960regression}  and its generalizations are non-experimental methods of causal inference in observational studies. In a typical regression discontinuity design,  the assignment of a treatment is determined according to whether a running variable  is greater or less than   a known cutoff while the outcome of interest  is supposed to be a smooth function of  the running variable except at the cutoff. The abrupt change in the probability of receiving the treatment can be exploited to infer the local causal effect around the cutoff. For general overviews, see \cite{imbens2008regression}, \cite{lee2010regression}, \cite{ce_regression_aie},  \cite{cattaneo2018practical1, cattaneo2018practical2}, and \cite{cattaneo2022}.

We develop in this study a family of estimators for Regression Discontinuity (RD) and Regression Kink (RK) under sharp and fuzzy designs. The proposed estimators are based on Gaussian Process (GP) regression/classification, a  probabilistic modeling approach that has gained popularity in machine learning and artificial intelligence community \citep{rasmussen2006}.  This nonparametric Bayesian approach places a GP prior on the latent function that underlays the outcome of interest. Modeler's knowledge regarding the underlying relationship is encoded through the mean and variance functions of the GP and inference is conducted according to Bayes' rule. 

Unlike methods that focus on the conditional mean of the outcome immediately below and above the cutoff (e.g., the local polynomial estimators), the proposed GP methods estimate the underlying functional relationship and rely on the resultant predictive distributions at the cutoff to infer various treatment effects of interest. One key advantage GP methods enjoy over other machine learning methods (such as neural network and random forest) is its automatic Uncertainty Quantification (UQ). Probabilistic quantities such as credibility intervals are readily obtained from the predictive distributions. Another appeal of GP methods is that its gradient remains a Gaussian process, facilitating derivative estimation. This is particularly valuable for the regression kink estimation considered in this study. 
Furthermore, thanks to the likelihood principle inherent in Bayesian inference, the proposed method infers both RD and RK effects from a common generative model and does not require separate models/configurations for the RD and RK estimation.

The fuzzy RD/RK design arises when  treatment assignment differs from actual treatment take-up. The identification of the RD/RK effects under the fuzzy design requires accounting for the change in probability of treatment take-up at the cutoff. Estimating this probability is a classification problem in which the GP method excels. The GP classification is a generalization of GP regression;  it models a continuous latent function underlaying the observed discrete outcome. We develop a GP RD estimator for binary outcomes and use it in conjunction with the GP regression to formulate estimators for fuzzy RD/RK effects.  

The essence of GP regressions is the covariance function that governs the sample path of a Gaussian process. We use the squared exponential covariance function in our GP regressions. It can be shown to be equivalent to a basis function expansion with an infinite number of radial basis functions and therefore rather expressive. It is, however, stationary such that the covariance between two inputs depends only on their distance. Consequently it may be inadequate if the underlying relationship is nonstationary. To tackle this potential limitation, we further develop a multi-layer hierarchical GP estimator. The first layer utilizes a Bayesian neural network that nonlinearly transforms the inputs into a latent feature space. The second layer utilizes the GP to map the intermediate outputs  to the observed responses. This estimator can be interpreted as a deep learning model with a multi-layer architecture. It combines the strength of neural network and GP regressions. Moreover using the GP estimator as the penultimate layer is amenable to  automatic uncertainty quantification and derivative estimation,  facilitating the estimation and inference of  RD/RK effects.

We conduct Monte Carlo simulations to assess the performance of the proposed methods. The results suggest that our estimators provide comparable and sometimes noticeably better performance relative to competing estimators. The hierarchical GP estimators improve upon the one-layer GP estimators substantially, demonstrating the merit of deep learning approach.  These overall patterns hold for the RD and RK estimations under both sharp and fuzzy designs, and across different data generating processes and sample sizes. For illustration, we apply the proposed methods to examine the electoral advantage to incumbency in the US House election \citep{lee2008randomized}. Lastly we present a further extension to accommodate covariate adjustment.

\section{Preliminaries}

\subsection{Regression-discontinuity  and regression-kink designs}
In this section, we provide a brief overview of regression discontinuity/kink designs; the main purpose is to introduce the various estimands that are considered in this study. There is an extensive and still growing literature on these topics;  readers are referred to \cite{imbens2008regression}, \cite{lee2010regression}, \cite{ce_regression_aie},  \cite{cattaneo2018practical1, cattaneo2018practical2}, and \cite{cattaneo2022} for general reviews.

%The RD design and its generalizations are non-experimental methods of causal inference in observational studies. In a typical RD design,  the assignment of a treatment is determined according to whether a running variable $ X  $ is greater or less than $ c $,  a known cutoff.  The outcome of interest $ Y$  is supposed to be a smooth function of  $ X $ except at the cutoff. 
Suppose that the assignment of a treatment  is determined according to whether a continuous running variable $ X $ is greater or less than $ c $, a known cutoff. It is assumed that units do not manipulate the running variable to position themselves above or below the cutoff.
The most common scenario is the Sharp RD (SRD)
design with a binary treatment, wherein all units above the cutoff receive the treatment and vice versa. The SRD treatment effect is defined as
\begin{equation}
	\tau_{\textsf{SRD}}=\lim_{x\downarrow c}\E[Y|X=x]-\lim_{x\uparrow c}\E[Y|X=x]\label{srd}
\end{equation}
\cite{hahn2001identification} established conditions under which causal effect is identified within this framework. 

More generally, suppose the outcome is given by $ Y=f(B,X,U) $, where $ B=B(X) $ is a policy/treatment parameter that depends on the running variable smoothly except at the cutoff and $ U $ is a stochastic error. The sharp regression kink (SRK) effect is defined as
\begin{equation}
	\tau_{\textsf{SRK}}=\frac{\lim_{x_0\downarrow c} \left.\frac{d\E[Y|X=x]}{dx}\right|_{x=x_0}-\lim_{x_0\uparrow c} \left.\frac{d\E[Y|X=x]}{dx}\right|_{x=x_0}}{\lim_{x_0\downarrow c} \left.\frac{dB(x)}{dx}\right|_{x=x_0}-\lim_{x_0\uparrow c} \left.\frac{dB(x)}{dx}\right|_{x=x_0}} \label{srk}
\end{equation}
Roughly speaking, the SRK parameter captures the local causal effect of a treatment on the derivative of $ f $ at the cutoff; see \cite{card2015inference} for a detailed analysis of this generalized RK design.

It is not uncommon in observational studies that actual treatment take-up differs from treatment assignment. In this case, the sharp RD/RK estimands introduced above reflect the effect of the treatment assignment rather than that of  the treatment take-up. Instead, the fuzzy RD/RK estimands identify the actual treatment effect by  exploiting the treatment assignment as an instrumental variable that affects the take-up probability at the cutoff. Define $ D=\1(\textnormal{treatment take-up} ) $. The fuzzy RD (FRD) and fuzzy RK (FRK) effects are given by
\begin{align*}
	\tau_{\textsf{FRD}}&=\frac{\lim_{x\downarrow c}\E[Y|X=x]-\lim_{x\uparrow c}\E[Y|X=x]}{\lim_{x\downarrow c}\E[D|X=x]-\lim_{x\uparrow c}\E[D|X=x]}\\
	\tau_{\textsf{FRK}}&=\frac{\lim_{x_0\downarrow c} \left.\frac{d\E[Y|X=x]}{dx}\right|_{x=x_0}-\lim_{x_0\uparrow c} \left.\frac{d\E[Y|X=x]}{dx}\right|_{x=x_0}}{\lim_{x_0\downarrow c} \left.\frac{d\E[B|X=x]}{dx}\right|_{x=x_0}-\lim_{x_0\uparrow c} \left.\frac{d\E[B|X=x]}{dx}\right|_{x=x_0}}
\end{align*}

Both parametric and nonparametric methods have been used in RD/RK estimations. Parametric estimations typically employ low order global polynomials; see e.g., \cite{van2002estimating} and \cite{arai2016optimal}. Nonparametric methods offer a flexible and robust alternative. Local polynomials, especially local linear estimators, are most commonly used due to its excellent boundary bias properties; see among others, \cite{hahn2001identification}, \cite{ludwig2007does}, \cite{imbens2012optimal}, and \cite{calonico2014robust}. Bayesian methods have also been used; see e.g. \cite{chib2016bayesian}, \cite{branson2019nonparametric} and \cite{chib2014nonparametric}.

\subsection{Gaussian process regression and classification}

In this sequel, we provide a brief introduction to the Gaussian Process methods.   Interested readers are referred to \cite{rasmussen2006} for an illuminating treatment of this topic.

Consider a sample $ \mathcal D $ consisting of responses $ \mb y=[y_1,\dots,y_N]^T $ and corresponding covariates/inputs $ \mb x=[x_1,\dots,x_N]^T  $. For simplicity, in this review we focus on the case of a single covariate; this is also in accordance with the typical RD/RK estimations wherein the running variable is the sole covariate. The response observations  are supposed to be conditionally independent given associated latent values $ \mb f=[f(x_1),\dots,f(x_N)]^T $. We place on the latent function $ f(\cdot) $ a GP prior with  mean function $ m(\cdot) $ and covariance function $ k(\cdot, \cdot) $. The observation model and the prior usually depend on some hyperparameters $ \theta_y$ and $ \theta_f $ respectively. 
Thus a typical GP model consists of the following components:
\begin{itemize}
	\item Observation model on $ \mb y $: $ \mb y|\mb f,\theta_y \sim \prod_{i=1}^N p(y_i|f(x_i),\theta_y) $
	\item GP prior on latent function $ f $: $ p(f (\mb x)|\theta_f ) \sim \mathcal{GP} (m(\mb x), k(\mb x,\mb x'|\theta_f)) $
	\item hyperprior on $ \theta=(\theta_f,\theta_y) $: $ \theta \sim p(\theta_f)p(\theta_y)  $
\end{itemize}
%Note that the observational model and the GP usually depend on different subsets of $ \theta $.  For notational simplicity, we do not make this distinction whenever there is not ambiguity.

The conditional posterior distribution of latent values $ \mb f $ given $  \theta $ takes the form
\begin{equation*}
	p(\mb f|\mathcal D, \theta)=\frac{p(\mb y|\mb f,\theta_y)p(\mb f|\mb x,\theta_f)}{\int p(\mb y|\mb f,\theta_y)p(\mb f|\mb x,\theta_f) d\mb f}
\end{equation*}
Integrating over $ \theta $ yields the marginal posterior
$
	p(\mb f|\mathcal D)=\int p(\mb f|\mathcal D,\theta)p(\theta)d\theta
$. Given a test point $ x_* $, let $ f_*=f(x_*) $ be its associated latent value. Denote by $ p(f_*|\mathcal D,\theta, x_*) $ the conditional predictive distribution of  $ f_* $. The corresponding conditional predictive distribution for the response is given by
$
p(y_*|\mathcal D, \theta, x_*)=\int p(y_*|f_*,\theta)p(f_*|\mathcal D,\theta, x_*)df_*
$. Its marginal distribution can be similarly obtained by integrating over $ \theta $.

A Gaussian process is a collection of random variables, any finite member of which are jointly Gaussian. It can be viewed as a generalization of the multivariate  Gaussian distribution to (infinite-dimensional) functional space. Throughout this study, we follow the convention of using $ \mathcal {GP}(\cdot,\cdot) $ and $ \mathcal N(\cdot,\cdot) $ to denote a GP and a multivariate Gaussian distribution respectively. The GP provides a powerful probabilistic framework for statistical learning. Consider first a Gaussian observation model $
	y_i=f(x_i)+u_i
$,
where $ u_i \sim \mathcal N (0,\sigma^2) $. The latent function is modeled as a GP with zero mean  and a covariance function $ k $ such that $ 	f(x) \sim \mathcal N(0, k(x,x')) $. There exists a large collection of covariance functions that are suitable to model relationship of various sorts; see Chapter 4 of \cite{rasmussen2006} for details. One popular choice  is the Squared Exponential (SE) covariance function:
\begin{equation}
	k_{\mathsf{ s}}(x,x')=\alpha^2 \exp{\left[- \frac{(x-x')^2}{2l^2}\right]}\label{se}
\end{equation}
where $\alpha^2$ is a scale parameter and the length scale $ l $ governs how fast the correlation between $ x $ and $ x' $ decreases with their distance.

Define $\mb K_{\mb x,\mb x'}=k(\mb x,\mb x'|\theta_f)$ and $ \mathcal I_N $ the $ N $-dimensional identity matrix. 
Note $\mb K_{x_*,\mb x} $ is a $ 1 \times N $ vector and $ K_{x_*,x_*} $ is a scalar. 
We then have
\begin{equation*}
	\left[ \begin{array} {c} \mb y \\f_* \end{array} \right] \sim
	\mathcal	N\left( {0}, \left[ \begin{array} {c c}   
		\mb K_{\mb x,\mb x}+\sigma^2 \mathcal I_N & \mb K_{\mb x,x_*}\\
		\mb K_{x_*,\mb x}&K_{x_*,x_*}
	\end{array} \right] \right)
\end{equation*}
The standard results on multivariate Gaussian distributions suggest that $ 	f_* | \mathcal{D},x_*,\theta \sim \mathcal N (m_*,v_*) 
 $, with the predictive mean and variance given by
\begin{equation}\label{fixed_predictive}
	\begin{aligned}
		m_* &=\E[f_*|\mathcal D,x_*,\theta]= {\mb K}_{x_*,\mb x} (\mb K_{\mb x,\mb x}+\sigma^2 \mathcal I_N)^{-1} {\mb y} \\
		v_* &=\V[f_*|\mathcal D,x_*,\theta]= K_{x_*,x_*} - {\mb K}_{x_*,\mb x} (\mb K_{\mb x, \mb x}+\sigma^2 \mathcal I_N)^{-1}{\mb K}_{\mb x,x_*}
	\end{aligned}
\end{equation}
It follows readily that 
$	y_*|\mathcal D,x_*,\theta \sim \mathcal N(m_*,v_*+\sigma^2)
$.

Since differentiation is a linear operator, the derivatives of a GP remains a GP. Assuming a twice-differentiable kernel, we define
\begin{equation*}
	\dot{k}(x,x')=\frac{\partial k(x,x')}{\partial x},\,	\ddot{k}(x,x')=\frac{\partial^2 k(x,x')}{\partial x\partial x'}
\end{equation*}
Let $\dot f_*=\partial f(x_*)/\partial x_*, \dot{\mb K}_{x_*,\mb x}=\dot k(x_*,\mb x|\theta_f)$ and $\ddot K_{x_*,x_*}=\ddot k(x_*,x_*|\theta_f) $. We can show that
$	\dot f_* |\mathcal{D},x_*,\theta	 \sim \mathcal N(\dot m_*,\dot v_* )
$ with
\begin{equation}\label{fixed_predictive_d}
	\begin{aligned}
		\dot m_* &  =\E[\dot f_*|\mathcal D,x_*,\theta]= {\dot{\mb K}}_{x_*,\mb x} (\mb  K_{\mb x,\mb x}+\sigma^2 \mathcal I_N)^{-1} {\mb y} \\
		\dot v_* &=\V[\dot f_*|\mathcal D,x_*,\theta]= \ddot K_{x_*,x_*} - {\dot{\mb K}}_{x_*,\mb x} (\mb K_{\mb x,\mb x}+\sigma^2 \mathcal I_N)^{-1}{\dot{\mb K}}_{\mb x,x_*}.
	\end{aligned}
\end{equation}
Below we shall exploit the differential GP  to develop GP estimators for RK effects.

For non-Gaussian observation models, the posterior distributions are generally not tractable.
The generalized GP regressions, in spirit close to the generalized linear models, are commonly used to tackle non-Gaussianity. 
%Both deterministic  and stochastic methods have been used to compute non-tractable posteriors. The former includes the Laplace approximation and the Expectation Propagation (EP), and the latter uses Markov Chain Monte Carlo (MCMC) methods. A prime example of generalized GP regression is the GP classification for categorical responses. We shall use it in our GP estimators of fuzzy RD/RK effect below. Although posterior distributions for generalized GP regressions are not tractable, 
Nonetheless,  GP's marginalization and conditionalization properties can still be exploited for prediction. Denote by $ \E[\mb f|\mathcal D,\theta] $ and $ \mathbbm V[\mb f|\mathcal D,\theta] $ the conditional mean and variance of the latent function associated with the conditional posterior $ p(\mb f|\mathcal D,\theta) $. The predictive mean and variance of $f( x_*) $ are given by
\begin{align}
	\E[f_*|\mathcal D,x_*,\theta]&=\mb K_{x_*,\mb x}\mb K_{\mb x,\mb x}^{-1}\E[\mb f|\mathcal D,\theta]\nonumber\\
	\V[f_*|\mathcal D,x_*,\theta]
	&=\left( K_{x_*,x_*}-\mb K_{x_*,\mb x}\mb K_{\mb x,\mb x}^{-1}\mb K_{\mb x,x_*}\right)+\mb K_{x_*,\mb x}\mb K_{\mb x,\mb x}^{-1}\V[\mb f|\mathcal D,\theta]\mb K_{\mb x,\mb x}^{-1}\mb K_{\mb x,x_*} \label{v*}\\
	&=K_{x_*,x_*}-\mb K_{x_*,\mb x}(\mb K_{\mb x,\mb x}^{-1}-\mb K_{\mb x,\mb x}^{-1} \V[\mb f|\mathcal D,\theta]\mb  K_{\mb x,\mb x}^{-1})\mb K_{\mb x,x_*}\nonumber
\end{align}
The first term on the right hand side of \eqref{v*} corresponds to the posterior variance of $ f_* $ conditional on a particular  $\mb  f $; the second term is due to the randomness of  $ \mb f $, which has a posterior variance $ \V[\mb f|\mathcal D,\theta ] $.

\begin{comment}
The marginal sample likelihood given the hyperparameters, $ p(\mathcal D|\theta)=\int p(\mb y|\mb f,\theta)p(\mb f|\mb x,\theta)d\mb f $, plays an important role in the inference of GP models. For Gaussian observation models, it admits the following analytical form
\begin{equation*}
	\log p(\mathcal D|\theta) =-\frac{1}{2}\mb y^T (\mb K_{\mb x,\mb x}+\sigma^2\mathcal I_N)^{-1}\mb y - \frac{1}{2}\log |\mb K_{\mb x,\mb x}+\sigma^2\mathcal I_N|-\frac{N}{2}\log 2\pi 	
\end{equation*}
The first term reflects the goodness-of-fit and the second term can be viewed as a penalty on model complexity, manifesting the intrinsic Bayesian Occam's Razor that  balances the fidelity to data and model simplicity. Although no general analytical expression exists for non-Gaussian models, the same underlying principle applies. 
\end{comment}

A variety of methods have been used for the inference of GP models. The marginal likelihood 
$ p(\mathcal D)=\int p(\mathcal D |\theta)p(\theta)d\theta $, sometimes referred to as the evidence, is indicative of the overall probability of the model. Since the computation of the marginal likelihood can be difficult, the Maximum a Posteriori (MAP) approach seeks the parameters that maximize its integrand:
$
	\hat{\theta}=\arg\max_{\theta} \log p(\mathcal D|\theta)+\log p(\theta)
$.
The posterior distribution of the latent function $ p(\mb f|\mathcal D) $ is then approximated by $ p(\mb f|\hat \theta) $. Since the posterior distribution of $ \theta $ is condensed to some point estimate, the MAP approach tends to underestimate a model's uncertainty. A second approach resorts to deterministic approximations such as the Laplace approximation, expectation propagation and variational methods. These methods approximate the posterior distribution with a simpler surrogate. Generally, the more flexible the approximation is, the better is the precision; however the computation cost usually increases with the degree of precision. A third possibility is to use Monte Carlo Markov Chain (MCMC) methods that construct a Markov chain whose stationary distribution coincides with the posterior distribution. One advantage of the MCMC methods is that it scales well in large parameter space as its rate of convergence is independent of the dimension of hyperparameters. This approach, however, can be computationally expensive. \cite{rasmussen2006} offer a detailed treatment of the first two approaches; \cite{gelman2013bayesian} provide a general overview of MCMC methods and a fully Bayesian treatment of GP models. The posterior consistency of GP regression and classification has been treated by \cite{ghosal2006posterior}, \cite{choi2007posterior}, and \cite{van2008rates,van2009adaptive}.

\section{GP models for sharp RD/RK designs}

In this section, we introduce GP estimators for RD/RK treatment effect under sharp and fuzzy designs. As mentioned above, a GP is characterized by its mean and covariance functions. Usually the zero mean function suffices. However when extrapolating out of sample, predictions of GP regression gravitate towards the mean function. Incorporating an explicit mean function may mitigate the `regressing to zero' tendency associated with a zero mean function in out of sample prediction. This is particularly desirable for RD estimation of local treatment effect at the cutoff, especially if the cutoff is not  covered by the sample range.

A common choice of mean function is a polynomial $ m(x)=\mb h^T(x)\beta $, where $  \mb h( x)=[1,x,\dots,x^S]^T$ and $ \beta=[\beta_{0},\beta_1,\dots,\beta_{S}]^T $. 
Within the Bayesian paradigm, we can absorb this mean function into a general composite covariance. With a Gaussian prior $ \beta \sim \mathcal N(0, \Sigma) $, $ m( x) \sim \mathcal N (0,\mb h^T(x) \Sigma   \mb h(x))$. Define 
\begin{equation*}
	k_{\textsf{p}}(x,x'|\Sigma)=\mb h^T(x) \Sigma \mb h(x')
\end{equation*}
This function turns out to be the so-called polynomial covariance function. One important appeal of GP regression is its expressiveness. One can not only choose from a large collection of covariance functions with varying features, but also construct new ones from the product and/or sum of covariances. In this study, we consider an additively composite covariance that is the sum of a polynomial covariance and a squared exponential covariance:
\begin{equation}
	k_{\textsf{p+s}}(x,x'|\Sigma,l,\alpha^2)=k_{\textsf{p}}(x,x'|\Sigma)+k_{\textsf{s}}(x,x'|l,\alpha^2)\label{kps}
\end{equation}

We can now proceed to RD estimations. We partition the sample into two subsets, denoting the subsample with $ x<c $ by $ \mathcal D_0=(\mb y_0,\mb x_0) $ and the rest by $ \mathcal D_1=(\mb y_1,\mb x_1) $, with respective subsample sizes $ N_0 $ and $ N_1 $. We first consider the sharp RD design. For $ j=0,1 $, we assume that
\begin{equation*}
	y_{i,j}=f_j(x_{i,j})+u_{i,j},i=1,\dots,N_j
\end{equation*}
where $ u_{i,j} \sim \mathcal N(0,\sigma_j^2)$. The latent function $ f_j $ has a GP prior with zero mean and composite covariance given by \eqref{kps}. We adopt the fully Bayesian approach for inference with independent prior distributions for $ \theta_j=(l_j,\alpha_j^2,\Sigma_j,\sigma_j^2) $. 

Denote by $ \tilde  \theta_{n,j}, n=1,\dots, \tilde N $, an MCMC sample of size $ \tilde N $ drawn from the posterior distribution $ p(\theta_j|\mathcal D_j) $. Define $ \mb K_{x,x'}(\theta)=k_{\textsf{p+s}}(x,x'|\theta) $. The predictive mean, conditional on $ \tilde \theta_{n,j} $, at the cutoff is given by
\begin{equation}
	\tilde	m_{n,j}= \mb K_{c,\mb x_j}(\tilde \theta_{n,j})(\mb K_{\mb x_j,\mb x_j}(\tilde \theta_{n,j})+\tilde \sigma_{n,j}^2 \mathcal I_{N_j})^{-1}\mb y_j \label{mij1}
\end{equation}
with conditional variance
\begin{equation*}
	\tilde	v_{n,j}=K_{c,c}(\tilde \theta_{n,j}) - {\mb K}_{c,\mb x_j}(\tilde \theta_{n,j}) (\mb K_{\mb x_j,\mb x_j}(\tilde \theta_{n,j})+\tilde\sigma^2_{n,j} \mathcal I_{N_j})^{-1}{\mb K}_{\mb x_j,c}(\tilde \theta_{n,j})
\end{equation*}
The predictive mean and variance are then calculated as
\begin{equation*}
	\hat \mu_j=\frac{1}{\tilde N}\sum_{n=1}^{\tilde N} \tilde m_{n,j},\quad 	\hat v_j=\frac{1}{\tilde N}\sum_{n=1}^{\tilde N}(\tilde m_{n,j}-\hat \mu_j)^2+\frac{1}{\tilde N}\sum_{n=1}^{\tilde N} \tilde v_{n,j}
\end{equation*}
where the first term of $ \hat v_j $ is the variance of conditional predictive mean and the second is the mean of conditional predictive variance. Finally the RD treatment effect is estimated by
\begin{equation*}
	\hat \tau_{\textsf{SRD}}=\hat \mu_1-\hat \mu_0
\end{equation*}
with posterior variance
\begin{equation*}
	\hat v_{\textsf{SRD}}=\hat v_1+\hat v_0
\end{equation*}

Next we consider the estimation of RK effect. Given a binary treatment, the denominator of the RK estimator \eqref{srk} equals one. 
Define $ \dot {  K}_{ x, x'}(\theta)=\dot k_{\textsf{p+s}}( x, x'|\theta) $ and $  \ddot K_{x,x'}(\theta)=\ddot k_{\textsf{p+s}}(x,x'|\theta) $. The predictive derivative, conditional on $ \tilde \theta_{n,j} $, at the cutoff is given by
\begin{equation*}
	\tilde {\dot	m}_{n,j}= \dot{\mb K}_{c,\mb x_j}(\tilde \theta_{n,j})(\mb K_{\mb x_j,\mb x_j}(\tilde \theta_{n,j})+\tilde \sigma_{n,j}^2 \mathcal I_{N_j})^{-1}\mb y_j
\end{equation*}
with conditional variance
\begin{equation*}
	\tilde	{\dot{v}}_{n,j}=\ddot K_{c,c}(\tilde \theta_{n,j}) - \dot{\mb K}_{c,\mb x_j}(\tilde \theta_{n,j}) (\mb K_{\mb x_j,\mb x_j}(\tilde \theta_{n,j})+\tilde\sigma^2_{n,j} \mathcal I_{N_j})^{-1}\dot{\mb K}_{\mb x_j,c}(\tilde \theta_{n,j})
\end{equation*}
Formulae for $ \dot{\mb K}_{c,\mb x_j} $ and $ \ddot{\mb K}_{c,c} $ are provided in Appendix A.
The predictive mean and variance of the derivative at the cutoff are  calculated as
\begin{equation*}
	\hat {\dot \mu}_j=\frac{1}{\tilde N}\sum_{n=1}^{\tilde N} \tilde {\dot m}_{n,j},\quad 	\hat {\dot v}_j=\frac{1}{\tilde N}\sum_{n=1}^{\tilde N}(\tilde {\dot m}_{n,j}-\hat {\dot \mu}_j)^2+\frac{1}{\tilde N}\sum_{n=1}^{\tilde N} \tilde {\dot v}_{n,j}
\end{equation*}
The  RK treatment effect is then estimated by
\begin{equation*}
	\hat \tau_{\textsf{SRK}}=\hat {\dot \mu}_1-\hat{\dot \mu}_0
\end{equation*}
with posterior variance
\begin{equation*}
	\hat v_{\textsf{SRK}}=\hat{\dot v}_1+\hat {\dot v}_0
\end{equation*}

\section{GP models for fuzzy RD/RK designs}

Under a fuzzy RD/RK design, the actual treatment take-up may differ from the treatment assignment. \cite{hahn2001identification} show that the fuzzy RD design is closely connected to an
instrumental variable  problem with a discrete instrument, and that the treatment effect of interest
is a version of the local average treatment effect or complier average treatment effect. As in the previous section, we start with the RD effect. The  estimation under fuzzy RD design entails estimating the ratio of  RD effects on the response and treatment take-up probability respectively. The former is given by the sharp RD estimator of the previous section, while the latter is based on the GP regression for binary responses, or GP classification.

%For a binary response $ D $, the GP classification assumes that the probability $ p(D=1|x)=\psi(f(x)) $, where $ f(x) $ has a GP prior and $ \psi $ is a sigmoid function such as the logistic or probit function. Note here the observation model $ \psi(f) $ is free of hyperparameters given the latent function $ f $. Given an iid sample $ \{D_i,x_i\}_{i=1}^N $, 

Under the fuzzy RD design, the probability of treatment take-up $p( D=1) $ is assumed to be a  smooth function of the running variable, except for a jump at the cutoff. Following the convention of the machine learning literature, we set $ D=-1 $ for no treatment take-up and $ D=1$ otherwise. 
Similarly to the SRD estimator, we model $ p(D=1) $ separately for the two subsamples $ \mathcal D_0=(\mb D_0,\mb x_0) $  and $ \mathcal D_1=(\mb D_1,\mb x_1) $ defined according to whether $ x \ge c $. For $ j=0,1 $ we assume that \begin{equation*}
	p(D_{i,j}=1|x_{i,j})=\psi(\gamma_j+D_{i,j}f_j(x_{i,j}))
\end{equation*}
where $ \psi $ is a sigmoid function such as the logistic or probit function,  $ \gamma_j $ is an offset coefficient, and $ f_j $ is a latent function with a GP prior.

The conditional marginal likelihood is given by
\begin{align*}
	\log p(\mathcal D_j|\theta_j)=&	\sum_{i=1}^{N_j} \log\psi(\gamma_j+D_{i,j}f_j(x_{i,j}))-\frac{1}{2}f_j(\mb x_j)^T\mb K_{\mb x_j,\mb x_j}^{-1}f_j(\mb x_j)\\
	&-\frac{1}{2}\log|\mb K_{\mb x_j,\mb x_j}|-\frac{N_j}{2}\log 2\pi
\end{align*}
The hyperparameter $ \theta_j $ includes $ \gamma_j $ for the observation model and those for the covariance of the latent function. For notational simplicity, the dependence of the covariance and latent function on the hyperparameters is suppressed.

Since the marginal likelihood $ p(D_j)=\int p(D_j|\theta_j)p(\theta_j)d\theta_j $ is not tractable, we use MCMC method for inference.
Let $ \tilde \theta_{n,j}, n=1,\dots,\tilde N $, be an MCMC sample of hyperparameters from the posterior distribution $ p(\theta_j|\mathcal D_j) $. For each $ \tilde \theta_{n,j} $, we  generate an $ N_j $-dimensional vector of  latent values according to
\begin{equation*}
	\tilde{\mb f}_{n,j} \sim \mathcal N(0,\mb K_{\mb x_j,\mb x_j}(\tilde \theta_{n,j}))
\end{equation*}
The predictive latent value at the cutoff is then given by
\begin{equation}
	\tilde	q_{n,j}= \mb K_{c,\mb x_j}(\tilde \theta_{n,j})\mb K^{-1}_{\mb x_j,\mb x_j}(\tilde \theta_{n,j})\tilde{\mb f}_{n,j} \label{mij2}
\end{equation}
To ensure numerical stability, a small positive constant is typically added to the diagonal of $ \mb K_{\mb x_j,\mb x_j} $ before taking its inverse. The corresponding predictive probability at the cutoff is then estimated by
$
\hat p_j=\frac{1}{\tilde N} \sum_{n=1}^{\tilde N}\tilde p_{n,j}
$, where  $
\tilde p_{n,j}= \psi(\tilde \gamma_{n,j}+\tilde q_{n,j})
$. Its variance is estimated by $ \hat v_{p,j}=\frac{1}{\tilde N_j}\sum_{n=1}^{\tilde N_j} (\tilde p_{n,j}-\hat p_j)^2 $.
The RD effect of treatment assignment on treatment take-up probability is calculated as
\begin{equation*}
	\hat \tau_{\mathsf{SRDP}}=\hat p_1-\hat p_0
\end{equation*}
with posterior variance 
\begin{equation*}
	\hat v_{\mathsf{SRDP}}=\hat v_{p,0}+\hat v_{p,1}
\end{equation*}

One can proceed to estimate the fuzzy RD effect with the ratio $ \hat \tau_{\mathsf{SRD} }/ \hat \tau_{\mathsf{SRDP} } $. This simple ratio estimator, however, is known to be biased and can be improved via a Jackknife refinement \citep{scott1981asymptotic,shao2012jackknife}. Denote by $  \hat\tau^{(-n)}_{\mathsf{SRD}} $ the sharp RD estimator on the response calculated with the $ n $-th MCMC sample left out, and $ \hat \tau^{(-n)}_{\mathsf{SRDP}} $ its counterpart on the treatment take-up probability.
The Jackknife FRD estimator is given by
\begin{equation*}
	\hat \tau_{\mathsf{FRD} }=\tilde N \frac{\hat\tau_{\mathsf{SRD} } }{\hat \tau_{\mathsf{SRDP} }}-\frac{\tilde N}{\tilde N-1}\sum_{n =1}^{\tilde N}\frac{\hat\tau^{(-n)}_{\mathsf{SRD} } }{\hat \tau^{(-n)}_{\mathsf{SRDP} }}
\end{equation*}
with variance
\begin{equation*}
	\hat v_{\mathsf{FRD}}=\frac{1}{\tilde N}\left(\frac{\hat v_{\mathsf{SRD}}}{\hat \tau^2_{\mathsf{SRDP}}}+\frac{\hat \tau^2_{\mathsf{SRD}}{\hat v_{\mathsf{SRDP}}}}{\hat \tau^4_{\mathsf{SRDP}}}  \right)
\end{equation*}

The estimator for the fuzzy RK effect is constructed analogously as follows:
\begin{equation*}
	\hat \tau_{\mathsf{FRK} }=\tilde N \frac{\hat\tau_{\mathsf{SRK} } }{\hat \tau_{\mathsf{SRDP} }}-\frac{\tilde N}{\tilde N-1}\sum_{n =1}^{\tilde N}\frac{\hat\tau^{(-n)}_{\mathsf{SRK} } }{\hat \tau^{(-n)}_{\mathsf{SRDP} }}
\end{equation*}
For fuzzy RD/RK design with a binary treatment, the FRK estimator shares with FRD the same  `denominator' estimator $ \hat \tau_{\mathsf{SRDP}} $. Thus its variance is estimated by
\begin{equation*}
	\hat v_{\mathsf{FRK}}=\frac{1}{\tilde N}\left(\frac{\hat v_{\mathsf{SRK}}}{\hat \tau^2_{\mathsf{SRDP}}}+\frac{\hat \tau^2_{\mathsf{SRK}}{\hat v_{\mathsf{SRDP}}}}{\hat \tau^4_{\mathsf{SRDP}}}  \right)
\end{equation*}

\section{Hierarchical GP models}

%\rt{see ``Being Bayesian, Even Just a Bit,
	%	Fixes Overconfidence in ReLU Networks'' and the argument for 'last layer Bayesian'}

%\rt{give RELU a shot; maybe not, as our method is not a gradient-based learning method}

To a large degree, the expressiveness of a GP model is determined by its covariance. The estimators introduced in the preceding sections employ the squared exponential covariance function. In spite of its  flexibility, it is stationary and can be inadequate if the underlying curve is non-stationary with  rapid variations in its slope/curvature.   There are two general ways to introduce non-stationarity to GP regressions \citep{rasmussen2006}. One is to make the parameters of the covariance function input dependent. However  it remains a challenge to maintain the positive-definiteness of the covariance with varying parameters. Another possibility is to introduce a non-linear transformation of the input $ x $, say $ g(x) $, and then construct a GP model in the $ g(x) $ space.

We adopt the second approach
and consider nonlinear transformations that are smooth and bounded, such as the logit, probit and tanh transformations. These sigmoid functions are widely used in neural network (NN) estimations, often referred to as activation functions. They have bounded derivatives and therefore provide desirable numerical stability, especially for RK estimations. %{The commonly used power transformation and Box-Cox transformation may not be suitable here because they  are unbounded and can cause exploding derivatives in the backpropagation of the training process.} 

\begin{figure}[h!]
	\centering
	\includegraphics[clip,trim=2.5cm 23cm 9cm 2.5cm, width=.6\textheight]{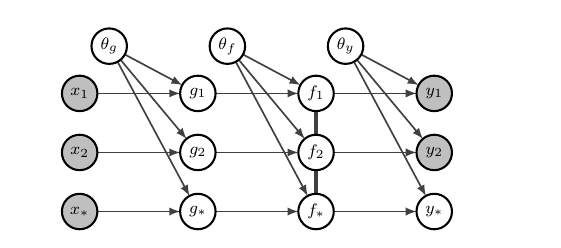}
	\caption{Graphical representation of proposed hierarchical GP model (the shaded nodes are observed, the rest are hidden; the third layer, modeled as a GP, is fully connected)}\label{fig:dgp}
\end{figure}

Given an activation function $ g $, we construct our first stage mapping 
$
g(x|\theta_g)=g(\lambda_0+\lambda_1 x)
$, where $ \theta_g=(\lambda_0,\lambda_1) $. 
We then apply the various GP models proposed in the preceding sections to $ g(x|\theta_g) $. For instance, the Gaussian observation model becomes $	y_i=f(g(x_i|\theta_g))+u_i
$. A Bayesian approach is adopted for the first stage as well. Thus the hyperparameters for this hierarchical GP model consist of $ \theta=(\theta_g,\theta_f,\theta_y) $ with prior distribution $ p(\theta_g)p(\theta_f)p(\theta_y) $. 
Figure \ref{fig:dgp} presents a graphical representation of the hierarchical GP model. This model can be interpreted as a four-layer deep learning model. The first and last layers are the inputs and outputs. The second layer is the NN transformation $ g(x) $ of the input $ x $ and the third layer constructs the GP latent variable based on $ g(x) $. Unlike the other layers, the third layer does not assume (conditional) independence  and instead models all units jointly as a Gaussian process.

Transformation of variables has had a long tradition in statistics. For instance, the power transformation and Box-Cox transformation have been customarily applied to non-Gaussian variables to reduce non-Gaussianity. In nonparametric estimations, the inputs are often transformed into  features (for instance splines, wavelets in series estimation and perceptrons in one-layer neural network) and the subsequent curve fitting is conducted on the feature space. Two key innovations  distinguish deep learning methods from these conventional methods \citep{murphy2012machine}. First, in contrast to a `shallow' model with a potentially large number of features, deep learning methods employ a multi-layer architecture, each layer nonlinearlly transforming its input into a slightly more abstract and composite representation. Second, many  conventional methods
that operate on the feature space can be characterized as un-supervised `feature engineering' wherein the construction of features and the subsequent fitting are disconnected. In contrast, deep learning methods  jointly train all layers. 
%For supervised learning tasks, deep learning methods eliminate feature engineering, by translating the data into compact intermediate representations akin to principal components, and derive layered structures that remove redundancy in representation.

The most popular deep learning models are multilayer neural networks \citep{Goodfellow-et-al-2016}. The proposed hierarchical  model can be characterized  a hybrid estimator that features an NN  layer followed by a GP layer. It is not uncommon that deep NN models 
feature thousands or even millions of tuning parameters and therefore can  be `data hungry'. Given the sometimes small samples for RD estimations, we opt for the more economic hybrid model with two hidden layers.   We prefer the GP approach to the NN as the penultimate layer for several reasons. First, it is probabilistic and offers automatic uncertainty quantification. Second, it is amenable to derivative estimations. Third, it is  more expressive than an NN  with a fixed number of nodes. As a matter of fact, the squared exponential covariance function can be obtained as the limit of basis function expansion with an infinite number of radial basis functions (see e.g., Chapter 4 of \cite{rasmussen2006}). 

%There are also deep GP models with multiple layers of Gaussian processes \citep{damianou2013deep}. Due to its computational complexity, the training of deep GP models typically resorts to approximate inference based on variational methods. This approach, however, is known to underestimate  model uncertainty and therefore not pursued in this study.

The marginal likelihood of the hierarchical GP model, conditional on $ \theta $, is given by
\begin{equation*}
	\log p(\mathcal D|\theta) =-\frac{1}{2}\mb y^T (\mb K_{g(\mb x),g(\mb x)}+\sigma^2\mathcal I_N)^{-1}\mb y - \frac{1}{2}\log |\mb K_{g(\mb x),g(\mb x)}+\sigma^2\mathcal I_N|-\frac{N}{2}\log 2\pi 	
\end{equation*}
and the unconditional marginal likelihood  $ p(\mathcal D)=\int p(\mathcal D|\theta)p(\theta_g)p(\theta_f)p(\theta_y)d\theta_g d\theta_f d \theta_y $.	Let $ \tilde  \theta_{n,j}=(\tilde \theta_{g,n,j},\tilde \theta_{f,n,j},\tilde \theta_{y,n,j}), n=1,\dots, \tilde N $, be an MCMC sample generated from the posterior distribution $ p(\theta_j|\mathcal D_j) $. 
Define $ {\mb K}_{g(x),g(x')}(\theta)=k_{\mathsf{p+s}}( g(\mb x|\theta_g),g(\mb x'|\theta_g)| \theta_{f}) $. The predictive mean, conditional on $ \tilde \theta_{n,j} $, at the cutoff is given by
\begin{equation}
	\tilde{m}_{n,j}(g)= \mb K_{g(c),g(\mb x_j)}(\tilde \theta_{n,j})(\mb K_{g(\mb x_j),g(\mb x_j)}(\tilde \theta_{n,j})+\tilde \sigma_{n,j}^2 \mathcal I_{N_j})^{-1}\mb y_j \label{mij2b}
\end{equation}
with conditional variance
\begin{align*}
	\tilde	v_{n,j}(g)=&K_{g(c),g(c)}(\tilde \theta_{n,j})\\
	& - {\mb K}_{g(c),g(\mb x_j)}(\tilde \theta_{n,j}) (\mb K_{g(\mb x_j),g(\mb x_j)}(\tilde \theta_{n,j})+\tilde\sigma^2_{n,j} \mathcal I_{N_j})^{-1}{\mb K}_{g(\mb x_j),g(c)}(\tilde \theta_{n,j})
\end{align*}

The predictive mean and variance at the cutoff are then calculated as
\begin{equation*}
	\hat{ \mu}_j(g)=\frac{1}{\tilde N}\sum_{n=1}^{\tilde N} \tilde m_{n,j}(g),\quad 
	\hat{ v}_j(g)=\frac{1}{\tilde N}\sum_{n=1}^{\tilde N}(\tilde m_{n,j}(g)-\hat{ \mu}_j(g))^2+\frac{1}{\tilde N}\sum_{n=1}^{\tilde N} \tilde v_{n,j}(g)
\end{equation*}
Finally the treatment effect is estimated by
\begin{equation*}
	\hat{\tau}_{\textsf{SRD}}(g)=\hat{ \mu}_1(g)-\hat{ \mu}_0(g)
\end{equation*}
with variance
\begin{equation*}
	\hat{v}_{\textsf{SRD}}(g)=\hat{v}_1(g)+\hat{v}_0(g)
\end{equation*}

Next consider the SRK effect.  Let $ \dot g(x|\theta)=\partial g(x|\theta_g)/\partial x $. Define $ \dot {  K}_{ g(x), g(x')}(\theta)=\dot k_{\textsf{p+s}}( g(x|\theta_g), g(x'|\theta_g)|\theta_f)\dot g(x|\theta_g) $ and $  \ddot K_{g(x),g(x')}(\theta)=\ddot k_{\textsf{p+s}}(g(x|\theta_g),g(x'|\theta_g)|\theta_f)\dot g(x|\theta_g)\dot g(x'|\theta_g) $.  The predictive derivative, conditional on $ \tilde \theta_{n,j} $, at the cutoff is given by
\begin{equation*}
	\tilde {\dot	m}_{n,j}(g)= \dot{\mb K}_{g(c),g(\mb x_j)}(\tilde \theta_{n,j})(\mb K_{g(\mb x_j),g(\mb x_j)}(\tilde \theta_{n,j})+\tilde \sigma_{n,j}^2 \mathcal I_{N_j})^{-1}\mb y_j
\end{equation*}
with conditional variance
\begin{align*}
	\tilde	{\dot{v}}_{n,j}(g)=&\ddot K_{g(c),g(c)}(\tilde \theta_{n,j})\\ 
	& - \dot{\mb K}_{g(c),g(\mb x_j)}(\tilde \theta_{n,j}) (\mb K_{g(\mb x_j),g(\mb x_j)}(\tilde \theta_{n,j})+\tilde\sigma^2_{n,j} \mathcal I_{N_j})^{-1}\dot{\mb K}_{g(\mb x_j),g(c)}(\tilde \theta_{n,j})
\end{align*}

The predictive mean derivative is then calculated as
\begin{equation*}
	\hat {\dot \mu}_j(g)=\frac{1}{\tilde N}\sum_{n=1}^{\tilde N} \tilde {\dot m}_{n,j}(g)
\end{equation*}
with predictive variance
\begin{equation*}
	\hat {\dot v}_j(g)=\frac{1}{\tilde N}\sum_{n=1}^{\tilde N}(\tilde {\dot m}_{n,j}(g)-\hat {\dot \mu}_j(g))^2+\frac{1}{\tilde N}\sum_{n=1}^{\tilde N} \tilde {\dot v}_{n,j}(g)
\end{equation*}
The SRK treatment effect is estimated by
\begin{equation*}
	\hat \tau_{\textsf{SRK}}(g)=\hat {\dot \mu}_1(g)-\hat{\dot \mu}_0(g)
\end{equation*}
with posterior variance
\begin{equation*}
	\hat v_{\textsf{SRK}}(g)=\hat{\dot v}_1(g)+\hat {\dot v}_0(g)
\end{equation*}
The hierarchical GP models  under the fuzzy 
RD and RK designs are constructed analogously. For brevity, the details are omitted.

\section{Numerical performance}

\subsection{Monte Carlo simulations}
We use Monte Carlo simulations to assess the numerical performance of the proposed methods. 
We examine the following data generating process that has been studied in various previous studies:
$
	y_i=f(x_i)+u_i
$, where $ x_i \sim 2\textnormal{Beta}(2,4)-1 $  and $ u_i \sim \mathcal N(0,0.1295^2) $.
We set the cutoff $ c=0 $ and consider the following mean functions:
\begin{itemize}
	
	\item DGP1: $ f(x)=0.42+0.1\mathbb I (x\ge 0)+0.84x-3x^2+7.99x^3-9.01x^4+3.56x^5 $
	\item DGP2: $f(x)=  (3.71+2.30x+3.28x^2+1.45x^3+0.23x^4+0.03x^5) \mathbb I(x<0) +(0.26+18.49x-54.18x^2+74.30x^3-45.02x^4+9.83x^5)\mathbb I(x\ge 0)$
	\item DGP3: $ f(x)=x^3 $
\end{itemize}
These designs feature different combinations of RD/RK effects. DGP1 has a moderate RD effect and no RK effect, DGP2 has sizable RD and RK effects, and DGP3 has neither RD nor RK effect. For the fuzzy RD/RK designs, we follow \cite{arai2016optimal} and set the treatment take-up probability
$ p(D=1|x<0)=\textnormal{Probit}(x|-1.28,1) $
and $ p(D=1|x\ge 0)=\textnormal{Probit}(x|1.28,1) $. This leads to a jump  in the treatment take-up probability of size 0.8 at the cutoff. We consider two sample sizes $ N=300$ and  500,  and repeat each experiment 300 times. 

We employ a fully Bayesian approach for the  inference of the proposed GP models. We denote the one- and two-layer GP models by GP1 and GP2 respectively. A half-normal $ \mathcal N^+(0,5^2) $ prior is placed on $ l,\alpha,\sigma $; a normal $ \mathcal N(0,5^2) $ prior is placed on the bias parameter $ \gamma $ of the GP classification model and the  parameters for the first layer of the hierarchical GP model $ \lambda_{0},\lambda_{1} $. The prior for the coefficients of the quadratic mean basis functions is set to be $ \mathcal N(0,100^2) $, yielding $ \Sigma=100^2 \mathcal I_3 $ in the corresponding polynomial covariance. These weakly informative priors have been customarily  used in previous studies. For our simulations, we use the {Stan} program that implements the Hamiltonian MCMC method \citep{neal2011mcmc}. We run four parallel Markov Chains; the number of MCMC draws is set to be 1000 with an equal number of warm-ups. To speed up computation,  we include in the estimation only observations with $ x $  no farther than twice of the Silverman's rule-of-thumb bandwidth from the cutoff. We use this commonly-used bandwidth mainly for convenience and twicing the bandwidth makes the thresholding rather generous.

For comparison, we also calculate the local linear (LL) estimators implemented in the R package `rdrobust'. We consider two versions of the LL estimators. LL1 uses the MSE-optimal bandwidth, and  LL2 is a robust alternative with bias-adjustment and coverage-optimal bandwidth; see \cite{calonico2014robust} for details. Note that different bandwidths are used for the LL estimation of RD and RK effects. In contrast, GP models obey the likelihood principle such that given the prior, all inferences and predictions depend only on the likelihood function. Therefore one common model is used for  the RD and RK estimations.

We summarize in Table \ref{tab:srdk} the estimation results for the SRD and SRK effects, reporting the average absolute bias, root mean squared error (RMSE), average coverage probability of 95\% confidence/credibility intervals and average interval length (IL). 
For each category, the estimator with the lowest bias, RMSE, IL or coverage probability closest to the nominal level is highlighted with the boldface font. In terms of both absolute bias and RMSE, the hierarchical GP estimator (GP2) dominates across the board, sometimes by considerable margins. GP2 noticeably improves upon the single-layer GP1. For instance for the RD effect, the average ratio of the RMSE between GP2 and GP1 across the three experiments is 0.69 for $ N=300 $ and 0.72 for $ N=500 $. Their counterparts for the RK effect are 0.61 and 0.60 respectively.
Mixed results are obtained for coverage and interval length.  LL2 and GP1 generally perform better in terms of coverage probability. As is reported in previous studies, LL2 improves upon LL1 in coverage at the expense of slightly wider intervals. It is also worth noting that the  GP estimators tend to have shorter intervals than the LL estimators.

To investigate the sensitivity of estimation results to the specification of tuning parameters, we experiment with some alternative estimation configurations. We report in Appendix B some additional estimation results under a range of prior distributions. These experiments suggest that our estimation results are rather stable with respect to these alternatives.

\begin{table}[h!]
	\centering
	\footnotesize
	\caption{SRD/SRK estimation results; top panel:  $ N=300 $, bottom panel: $ N=500 $}
	\label{tab:srdk}
	\begin{tabular}{clcccccccc}
		\hline
		&&\multicolumn{4}{c}{SRD}&\multicolumn{4}{c}{SRK}\\\cline{3-10}
		&			&	LL1		&LL2	&GP1	&GP2	&LL1	&LL2	&GP1	&GP2	\\
		%	&			&	RD		&RD		&RD		&RD		&RK		&RK		&RK		&RK     \\
		\hline
		DGP1&	$ |\textnormal{Bias}| $	&	0.065	&0.076	&0.069	&\textbf{0.046}	&1.363	&1.791	&1.276	&\textbf{1.161}  \\
		&	RMSE	&	0.084	&0.097	&0.088	&\textbf{0.058}	&1.752	&2.423	&1.629	&\textbf{1.287}  \\
		&	Coverage&	0.903	&0.913	&\textbf{0.970}	&0.983	&0.863	&\textbf{0.943}	&0.970	&0.743  \\
		&	IL		&	0.302	&0.356	&0.365	&\textbf{0.260}	&5.363	&8.298	&7.405	&\textbf{3.798}  \\\cline{3-10}
		DGP2&	$|\textnormal{Bias}| $	&	0.100	&0.087	&0.075	&\textbf{0.072}	&3.771	&2.594	&2.101	&\textbf{1.666}  \\
		&	RMSE	&	0.123	&0.110	&0.094	&\textbf{0.090}	&4.267	&3.119	&2.506	&\textbf{2.095}  \\
		&	Coverage&	0.890	&\textbf{0.960}	&0.963	&\textbf{0.967}	&0.580	&\textbf{0.950}	&0.853	&0.990  \\
		&	IL		&	0.453	&0.502	&\textbf{0.368}	&0.388	&9.865	&12.982	&\textbf{7.618}	&10.764 \\\cline{3-10}
		DGP3&	$|\textnormal{Bias}| $	&	0.061	&0.071	&0.068	&\textbf{0.032}	&0.881	&1.397	&1.254	&\textbf{0.212}  \\
		&	RMSE	&	0.076	&0.089	&0.087	&\textbf{0.041}	&1.274	&1.978	&1.607	&\textbf{0.305}  \\
		&	Coverage&	0.907	&0.917	&\textbf{0.967}	&0.987	&\textbf{0.953}	&0.963	&0.970	&0.990  \\
		&	IL		&	0.274	&0.326	&0.364	&\textbf{0.198}	&4.149	&6.660	&7.401	&\textbf{1.917}  \\\hline
		DGP1&	$|\textnormal{Bias}| $	&	0.050	&0.057	&0.053	&\textbf{0.036}	&1.112	&1.366	&1.167	&\textbf{1.053} \\
		&	RMSE	&	0.066	&0.075	&0.067	&\textbf{0.045}	&1.353	&1.772	&1.475	&\textbf{1.186} \\
		&	Coverage&	0.903	&0.913	&\textbf{0.963}	&0.980	&0.880	&\textbf{0.950}	&0.970	&0.730 \\
		&	IL		&	0.237	&0.277	&0.282	&\textbf{0.207}	&4.381	&6.667	&6.297	&\textbf{3.310} \\\cline{3-10}
		DGP2&	$|\textnormal{Bias}| $	&	0.074	&0.067	&0.056	&\textbf{0.054}	&3.311	&2.149	&1.757	&\textbf{1.410} \\
		&	RMSE	&	0.094	&0.087	&0.071	&\textbf{0.070}	&3.760	&2.660	&2.160	&\textbf{1.792} \\
		&	Coverage&	0.880	&0.920	&\textbf{0.947}	&0.960	&0.590	&\textbf{0.937}	&0.837	&0.983 \\
		&	IL		&	0.319	&0.348	&\textbf{0.283}	&0.294	&8.037	&10.217	&\textbf{6.419}	&8.739 \\\cline{3-10}
		DGP3&	$|\textnormal{Bias}| $	&	0.045	&0.052	&0.053	&\textbf{0.026}	&0.654	&1.012	&1.157	&\textbf{0.179} \\
		&	RMSE	&	0.059	&0.068	&0.067	&\textbf{0.033}	&0.926	&1.401	&1.462	&\textbf{0.244} \\
		&	Coverage&	0.920	&0.933	&\textbf{0.960}	&0.990	&0.940	&\textbf{0.957}	&0.970	&1.000 \\
		&	IL		&	0.205	&0.243	&0.282	&\textbf{0.154}	&2.981	&4.750	&6.297	&\textbf{1.595} \\
		\hline
	\end{tabular}
\end{table}

We next examine estimation results under fuzzy designs.
The first stage of our  FRD/FRK estimators
uses the GP classification to estimate the jump in probability of treatment take-up at the cutoff. The GP estimators perform well for this task (see Appendix C for details).  
Table \ref{tab:frdk} reports the estimation results of the fuzzy RD/RK effects. The overall pattern is similar to those under the sharp design. GP2 again dominates in terms of estimation precision, registering the smallest absolute bias and RMSE 
in all but one instances. It also tends to have the shortest confidence interval, while LL2 and GP1 generally excel in coverage probability. Similarly to the cases under a sharp design, GP2 considerably improves upon GP1. The average ratios of the RMSE between GP2 and GP1 for the RD effects are 0.64 for both sample sizes; for the RK effects, they are 
0.60 for $ N=300 $ and 0.58 for $ N=500 $.
\begin{table}[h!]
	\centering
	\footnotesize
	\caption{FRD/FRK estimation results; top panel:  $ N=300 $, bottom panel: $ N=500 $}\label{tab:frdk}
	\begin{tabular}{clcccccccc}
		\hline
		&&\multicolumn{4}{c}{FRD}&\multicolumn{4}{c}{FRK}\\\cline{3-10}
		&			&	LL1		&LL2	&GP1	&GP2	&LL1	&LL2	&GP1	&GP2	\\
		%&			&	RD		&RD		&RD	&RD		&RK		&RK		&RK		&RK     \\
		\hline	
		DGP1&	$|\textnormal{Bias}| $	&	0.098	&0.123	&0.085	&\textbf{0.058}	&\textbf{1.124}	&2.068	&1.782	&1.377 \\
		&	RMSE	&	0.162	&0.187	&0.109	&\textbf{0.072}	&1.967	&5.814	&2.214	&\textbf{1.540} \\
		&	Coverage&	0.909	&0.909	&\textbf{0.951}	&0.972	&0.972	&\textbf{0.962}	&0.972	&0.875 \\
		&	IL		&	0.411	&0.486	&0.447	&\textbf{0.321}	&18.292	&30.139	&9.186	&\textbf{4.867} \\\cline{3-10}
		DGP2&	$|\textnormal{Bias}| $	&	0.597	&0.791	&0.395	&\textbf{0.316}	&18.681	&19.041	&6.225	&\textbf{6.105} \\
		&	RMSE	&	1.136	&1.471	&0.494	&\textbf{0.400}	&21.046	&22.916	&8.093	&\textbf{7.625} \\
		&	Coverage&	0.920	&\textbf{0.951}	&0.970	&1.000	&0.420	&0.561	&\textbf{0.932}	&0.864 \\
		&	IL		&	2.509	&2.955	&1.937	&\textbf{1.852}	&71.413	&120.028&\textbf{26.794}	&27.085 \\\cline{3-10}
		DGP3&	$|\textnormal{Bias}| $	&	0.093	&0.108	&0.082	&\textbf{0.037}	&0.788	&1.185	&1.740	&\textbf{0.244} \\
		&	RMSE	&	0.167	&0.178	&0.104	&\textbf{0.046}	&1.400	&2.882	&2.162	&\textbf{0.364} \\
		&	Coverage&	\textbf{0.934}	&0.931	&0.969	&0.997	&0.972	&0.972	&\textbf{0.962}	&1.000 \\
		&	IL		&	0.366	&0.433	&0.443	&\textbf{0.235}	&14.713	&23.908	&9.126	&\textbf{2.438} \\\hline
		DGP1&	$|\textnormal{Bias}| $	&	0.066	&0.081	&0.071	&\textbf{0.051}	&1.240	&2.137	&1.538	&\textbf{1.221} \\
		&	RMSE	&	0.096	&0.116	&0.090	&\textbf{0.061}	&2.154	&5.086	&1.947	&\textbf{1.383} \\
		&	Coverage&	0.929	&0.908	&\textbf{0.954}	&0.982	&0.993	&0.972	&\textbf{0.958}	&0.852 \\
		&	IL		&	0.288	&0.338	&0.353	&\textbf{0.259}	&21.869	&35.813	&8.107	&\textbf{4.246} \\\cline{3-10}
		DGP2&	$|\textnormal{Bias}| $	&	0.355	&0.531	&0.351	&\textbf{0.279}	&18.315	&18.604	&5.918	&\textbf{5.582} \\
		&	RMSE	&	0.595	&0.788	&0.433	&\textbf{0.345}	&21.561	&21.984	&7.529	&\textbf{6.637} \\
		&	Coverage&	0.976	&0.984	&\textbf{0.951}	&0.992	&0.565	&0.671	&\textbf{0.927}	&0.858 \\
		&	IL		&	1.542	&1.830	&1.607	&\textbf{1.535}	&85.993	&142.616&24.140	&\textbf{23.939} \\\cline{3-10}
		DGP3&	$|\textnormal{Bias}| $	&	0.060	&0.071	&0.071	&\textbf{0.032}	&0.857	&1.441	&1.437	&\textbf{0.210} \\
		&	RMSE	&	0.092	&0.104	&0.088	&\textbf{0.039}	&1.773	&5.101	&1.825	&\textbf{0.286} \\
		&	Coverage&	0.934	&0.920	&\textbf{0.955}	&0.993	&0.997	&0.997	&\textbf{0.965}	&1.000 \\
		&	IL		&	0.255	&0.300	&0.349	&\textbf{0.189}	&18.396	&30.100	&7.996	&\textbf{2.007} \\\hline
	\end{tabular}
\end{table}

\subsection{Empirical application}
We apply the proposed methods to the US House election data studied by \cite{lee2008randomized}. This study investigates the electoral advantage to incumbency in elections to the United States House of
Representatives. Many factors contribute to successful political campaigns and elections. \cite{lee2008randomized} argued that districts where a party’s candidate just barely won an election and hence barely became the incumbent are likely to be comparable in all other ways to districts where the
party’s candidate just barely lost the election. Therefore differences in the electoral success between these two groups in
the next election can be exploited to identify the causal party incumbency advantage.

In this study, the treatment corresponds to winning the previous election, and the running variable  corresponds to the margin of victory of a Democratic
candidate (with the cutoff set at zero). The outcome of interest is the Democratic vote share in the following election. Elections with the previous winning margin between positive and negative 25\% are used in the estimation. 
To avoid potential censoring issues, we also omit extreme observations with zero or one hundred percent Democratic votes. The number of retained observations is 2681.

We estimate the SRD/SRK effects of incumbency on vote share. In addition,  we also estimate the treatment effect on the probability of getting more than 50\% of votes, which guarantees a victory. The results are reported in Table \ref{tab:application}. All four estimators point to a significant RD effect at the level of 5-7 percentage point increase in vote share. GP2 estimate is closer to the two LL estimates and improves upon GP1 in terms of precision. Although all estimators suggest a positive RK effect, none of them is statistically significant. LL1 and GP2 estimates are close in magnitude and precision. 
The estimation results on the probability of winning are reported in the third row. As explained in the previous section, GP1 and GP2 generate virtually identical results for classification tasks, thus we focus on GP1 in the estimation of winning probability. All estimators suggest a positive RD effect about 50\% increase in the probability of winning the election. The GP estimate has a considerably smaller standard error than the two LL estimates.

\begin{table}[h!]
	\centering
	\small
	\caption{Estimated SRD/SRK effect on vote share (rows one and two) and SRD effect on winning probability (row three)}
	\label{tab:application}
	\begin{tabular}{lcccccccc}
		\hline
		&\multicolumn{2}{c}{LL1}&\multicolumn{2}{c}{LL2}&\multicolumn{2}{c}{GP1}&\multicolumn{2}{c}{GP2}\\\cline{2-9}
		&$ \hat \tau $&se$ (\hat \tau) $&$ \hat \tau $&se$ (\hat \tau) $&$ \hat \tau $&se$ (\hat \tau) $&$ \hat \tau $&se$ (\hat \tau) $ \\\hline
		
		SRD &5.81&1.18&5.89&1.39&7.01&2.22&6.23&1.65\\
		SRK &33.44&49.09&69.11&77.73&54.53&166.60&24.51&52.64\\
		SRDP &0.53&0.16&0.49&0.19&0.48&0.04&---&---\\
		\hline
	\end{tabular}
\end{table}

\begin{figure}[h!]
	\centering
	\includegraphics[width=.49\textwidth]{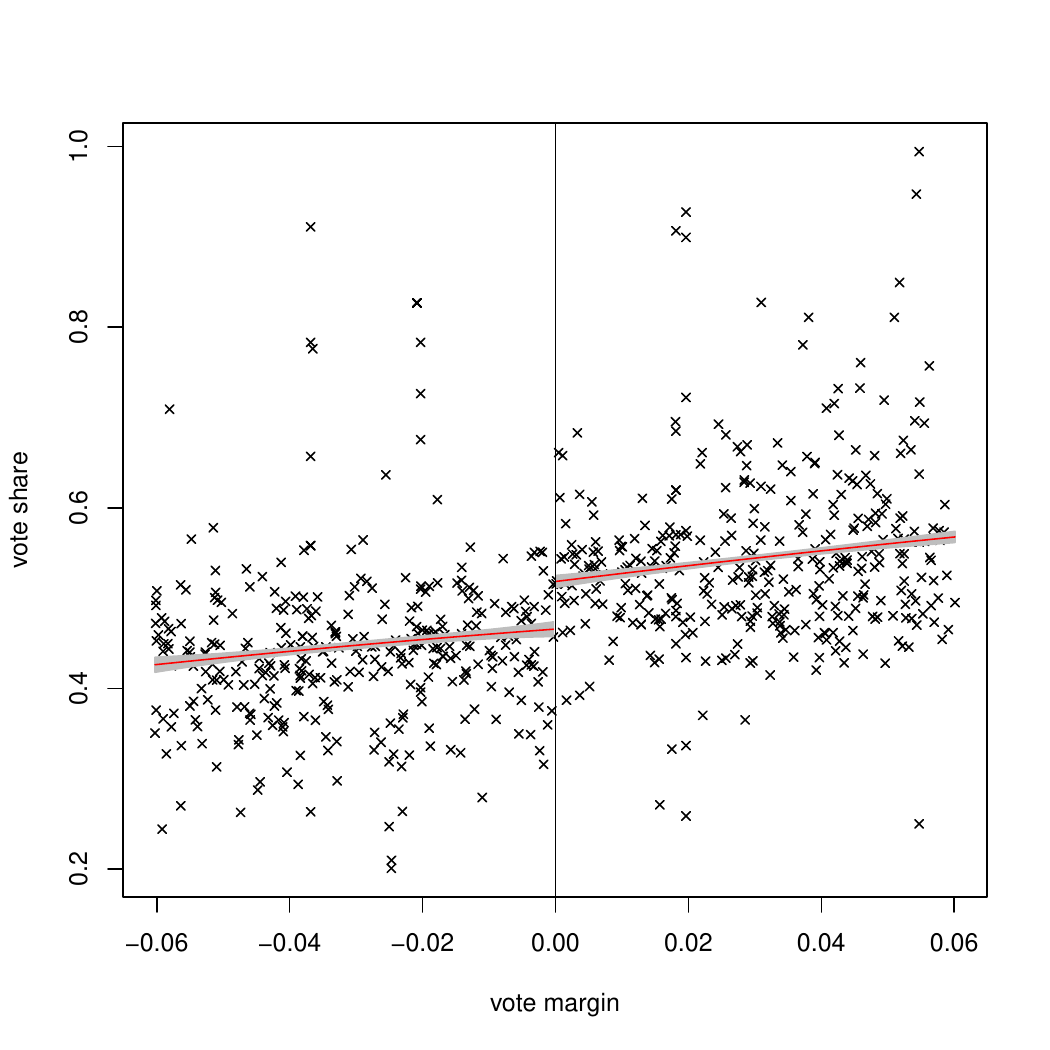}	\includegraphics[width=.49\textwidth]{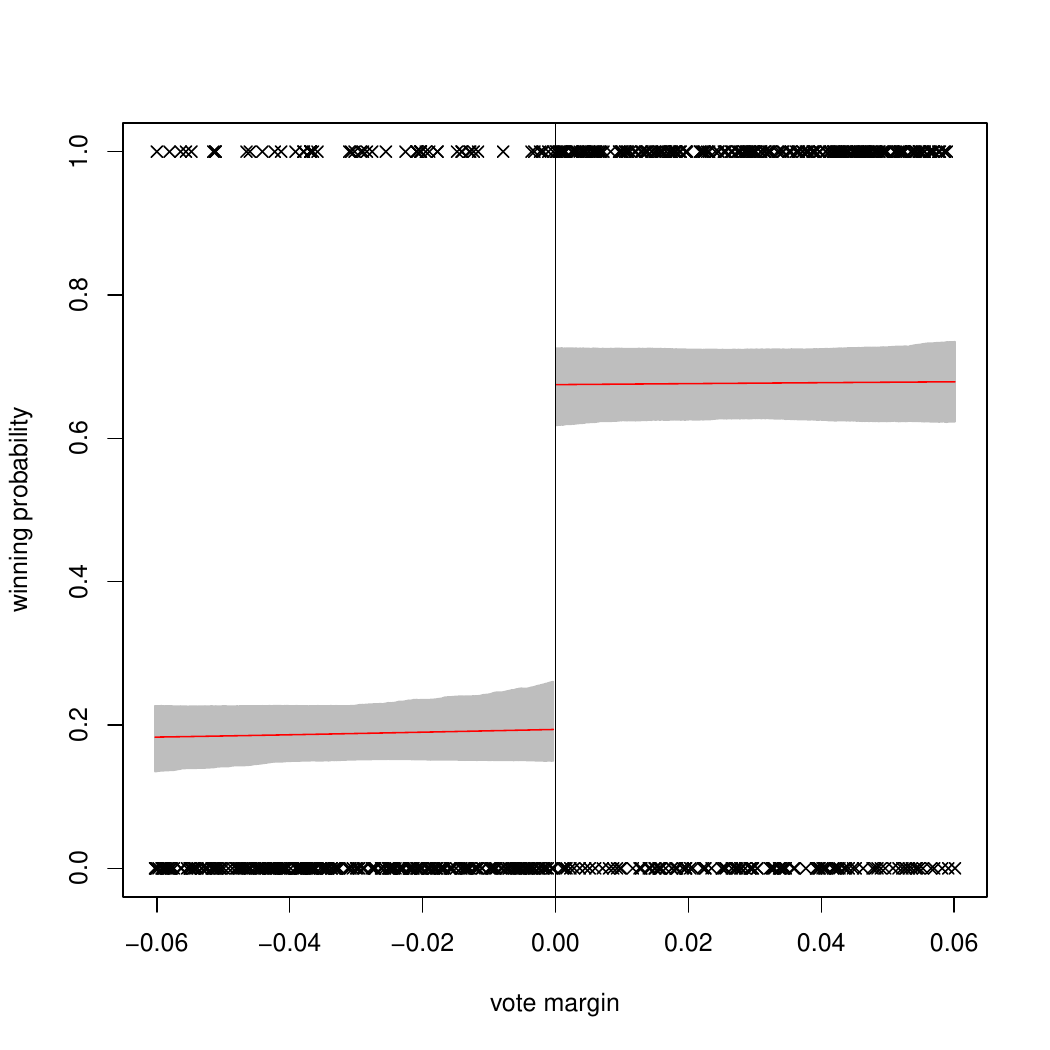}
	\caption{Estimated latent function for vote share (left) and winning probability (right); shaded band indicates 95\% credible interval of the predictive mean}\label{fig:vote}
\end{figure}

The predictive means and their 95\% credible intervals of the estimated latent functions for vote share (based on GP2) and winning probability (based on GP1) are  plotted in Figure \ref{fig:vote}. Both indicate unmistakable effects of incumbency advantage. The credible intervals for the winning probability estimation is noticeably wider; this can be attributed to information loss caused by the reduction of a continuous response variable to a binary one.

\section{Conclusion and further extension for covariate adjustment}

We propose a family of Gaussian Process estimators for causal inference exploiting Regression Discontinuity/Kink (RD/RK) under sharp and fuzzy designs.   The GP methods are powerful probabilistic modeling approaches that are advantageous in terms of derivative estimation and uncertainty quantification, facilitating RK estimation and inference of RD/RK models. These estimators are extended to  hierarchical GP models with an intermediate Bayesian neural network layer. This estimator  can be characterized as a hybrid deep learning model. Monte Carlo simulations show that our estimators perform comparably to and sometimes better than competing estimators in terms of precision, coverage and interval length. The hierarchical GP models improve up one-layer GP models substantially. 

Lastly we briefly present a further extension to incorporate covariates. Covariate adjustment based on pre-intervention measures may allow for efficiency gains or the evaluation of treatment effect heterogeneity. Recent studies have explored the inclusion of covariates in RD estimations; see a recent review by \cite{cattaneo2021covariate} and references therein. It is relatively straightforward to include covariates in GP regressions. 
 \cite{calonico2019regression} suggest that in continuity-based RD estimation, covariates  should be entered in an additive separable manner. Suppose in addition to the running variable $ x $, one is to incorporate a vector of covariates $ z $. This can be modeled by a Gaussian Process with  an additive composite covariance $  k_1(x,x')+k_2(z,z') $, where  $ k_1 $ and $ k_2 $ are generic covariance functions. The hyperparameters of these two covariances are jointly learned during the estimation. The influence of the covariates is then profiled out and inference on RD effect then proceeds in the same manner as the `unadjusted' estimates dicussed in the previous sections. 
 As an illustration, we apply this covariate-adjusted estimator to the same US House election data. To save space, we focus on the one-layer GP estimator of sharp RD effect on vote share. The estimated coefficient and standard deviation are respectively 6.49 and 2.19, which are close to the `unadjusted' estimates of 7.01 and 2.22. A detailed investigation of covariate-adjusted estimation is beyond the scope of this study. For completeness, necessary details to implement the covariate-adjusted RD estimator are provided in  Appendix D. 
%%%%%%%%%%%%%%%%%%%%%%%%%%%%%%%%%%%%%%%%%%%%%%
%% Single Appendix:            %%
%%%%%%%%%%%%%%%%%%%%%%%%%%%%%%%%%%%%%%%%%%%%%%
%\begin{appendix}
%\section*{???} %% if no title is needed, leave empty \section*{}.
%\end{appendix}
%%%%%%%%%%%%%%%%%%%%%%%%%%%%%%%%%%%%%%%%%%%%%%
%% Multiple Appendixes:        %%
%%%%%%%%%%%%%%%%%%%%%%%%%%%%%%%%%%%%%%%%%%%%%%
%\begin{appendix}
%\section{???}
%
%\section{???}
%\end{appendix}

\begin{appendix}
	\section*{Appendix}
	\subsection*{A. Posterior variance of differentiate GP}
	%Below we present the details in the construction of posterior variance of the proposed estimators.
	
	Recall that we use a composite covariance $ k_{\mathsf{p+s}}(x,x')=k_{\mathsf{p}}(x,x')+k_{\mathsf{s}}(x,x')$, where $ k_{\mathsf{p}}(x,x'|\Sigma)=\mb h^T(x)\Sigma \mb h(x') $ and $ k_{\mathsf{s}}(x,x'|l,\alpha^2)=\alpha^2 \exp(-\frac{(x-x')^2}{2l^2}) $.
	It follows that
	\begin{equation*}
		\dot k_{\mathsf{p}}(x,x'|\Sigma)=	\frac{\partial k_{\mathsf{p}}(x,x'|\Sigma)}{\partial x}= \frac{\partial \mb h^T(x)}{\partial x}\Sigma \mb h(x')
	\end{equation*}
	and
	\begin{equation*}
		\ddot k_{\mathsf{p}}(x,x'|\Sigma)=	\frac{\partial ^2 k_{\mathsf{p}}(x,x'|\Sigma)}{\partial x \partial x'}=\frac{\partial \mb h^T(x)}{\partial x}\Sigma \frac{\partial \mb h(x')}{\partial x'}
	\end{equation*}
	Next for the SE covariance, we have
	\begin{equation*}
		\dot k_{\mathsf{s}}(x,x'|l,\alpha^2)=\frac{\partial k_{\mathsf{s}}(x,x'|l,\alpha^2)}{\partial x}=-\frac{x-x'}{l^2}k_{\mathsf{s}}(x,x'|l,\alpha^2)
	\end{equation*}
	and
	\begin{equation*}
		\ddot k_{\mathsf{s}}(x,x'|l,\alpha^2)=\frac{\partial ^2 k_{\mathsf{s}}(x,x'|l,\alpha^2)}{\partial x \partial x'}=\frac{1}{l^2}k_{\mathsf{s}}(x,x'|l,\alpha^2)+\frac{(x-x')^2}{l^4}k_{\mathsf{s}}(x,x'|l,\alpha^2)
	\end{equation*}
	It follows that $
	\ddot k_{\mathsf{s}}(x,x|l,\alpha^2)=\frac{\alpha^2}{l^2}
	$.
	
	Gathering these results, we have
	\begin{align*}
		\dot {\mb K}_{c,\mb x}(\theta)&=\left.\frac{\partial \mb h^T(x)}{\partial x}\right|_{x=c}\Sigma \mb h(\mb x)-\frac{c-\mb x}{l^2}k_{\mathsf{s}}(c,\mb x|l,\alpha^2)\\
		\ddot { K}_{c,c}(\theta)&=\left.\frac{\partial \mb h^T(x)}{\partial x}\right|_{x=c}\Sigma \left.\frac{\partial \mb h(x)}{\partial x}\right|_{x=c}+\frac{\alpha^2}{l^2}
	\end{align*}

\subsection*{B. Estimation results under alternative configurations}
We have experimented with many aspects of the estimation configuration. The results are not sensitive to alternative specifications. It is infeasible to report all these results. Below we report experiments on alternative prior distributions  for the key tuning parameters $ l,\alpha, \sigma $ of the covariances, focusing on the  sharp RD/RK estimation of single-layer GP regressions. Instead of the half normal $\mathcal N^+(0,5^2) $ prior used in the simulations reported in the text, we consider two alternatives: 	$\mathcal N^+(0,3^2) $ and $ \mathcal N^+(0,10^2) $. The former is somewhat restrictive while the latter is highly non-informative. The results reported in Table A.1 clearly suggest that our estimation is not sensitive to the range of variation in  prior distributions. For $ N=500 $, the results obtained under different prior distributions are hardly distinguishable. This is consistent with the general expectation that in Bayesian analysis, the importance of the prior tends to diminish with the sample size.

\begin{table}[h!]
	\centering
	\footnotesize
\centering
\small
Table A.1	
SRD/SRK estimation results with prior distributions for $ l,\alpha,\rho $ at $ \mathcal N^+(0,a^2) $\\
Top panel: $ N=300 $; Bottom panel: $ N=500 $
\vskip .1in
%	\label{tab:srdk}
	\begin{tabular}{clcccccccc}
		\hline
		&&\multicolumn{3}{c}{SRD}&\multicolumn{3}{c}{SRK}\\\cline{3-8}
		&			&	$ a=3 $		&$ a=5 $&$ a=10 $	&$ a=3 $	&$ a=5 $	&$ a=10 $	\\
		\hline
		DGP1&	$ |\textnormal{Bias}| $	&	0.069	&0.069	&0.069	&1.281	&1.276	&1.272	  \\
		&	RMSE	&	0.088	&0.088	&0.087	&1.632	&1.629	&1.625	  \\
		&	Coverage&	0.970	&0.970	&0.967	&0.973	&0.970	&0.970	  \\
		&	IL		&	0.366	&0.365	&0.361	&8.008	&7.405	&7.252	 \\\cline{3-8}
		DGP2&	$ |\textnormal{Bias}| $	&	0.075	&0.069	&0.075	&2.088 &2.101	&2.117	  \\
&	RMSE	&	0.094	&0.088	&0.094	&2.497	&2.506	&2.522	  \\
&	Coverage&	0.957	&0.963	&0.950	&0.873	&0.853	&0.853	  \\
&	IL		&	0.369	&0.368	&0.363	&8.103	&7.618	&7.372	 \\\cline{3-8}
		DGP3&	$ |\textnormal{Bias}| $	&	0.068	&0.068	&0.068	&1.257	&1.254	&1.251	  \\
&	RMSE	&	0.087	&0.088	&0.087	&1.609	&1.607	&1.607	  \\
&	Coverage&	0.967	&0.967	&0.967	&0.977	&0.970	&0.967	  \\
&	IL		&	0.365	&0.364	&0.361	&7.665	&7.401	&7.264
\\\hline
DGP1&	$ |\textnormal{Bias}| $	&	0.053	&0.053	&0.053	&1.170	&1.167	&1.163	  \\
&	RMSE	&	0.067	&0.067	&0.067	&1.478	&1.475	&1.473	  \\
&	Coverage&	0.963	&0.963	&0.957	&0.973	&0.970	&0.967	  \\
&	IL		&	0.282	&0.282	&0.280	&6.621	&6.297	&6.189	 \\\cline{3-8}
DGP2&	$ |\textnormal{Bias}| $	&	0.056	&0.056	&0.053	&1.753 &1.757	&1.769	  \\
&	RMSE	&	0.071	&0.071	&0.067	&2.156	&2.160	&2.168	  \\
&	Coverage&	0.947	&0.947	&0.957	&0.843	&0.837	&0.833	  \\
&	IL		&	0.285	&0.283	&0.280	&6.719	&6.419	&6.278	 \\\cline{3-8}
DGP3&	$ |\textnormal{Bias}| $	&	0.053	&0.053	&0.053	&1.161	&1.157	&1.153	  \\
&	RMSE	&	0.067	&0.067	&0.067	&1.467	&1.462	&1.460	  \\
&	Coverage&	0.960	&0.960	&0.957	&0.977	&0.970	&0.967	  \\
&	IL		&	0.282	&0.282	&0.280	&6.544	&6.297	&6.236
\\\hline

	\end{tabular}
\end{table}

\subsection*{C. Estimation results on treatment take-up probability}
Our numerical experiments suggest little difference in the performance between GP1 and GP2 in the estimation of treatment take-up probability. We therefore use GP1 in our simulations due to its lower computation cost.
{Since the same procedure of treatment take-up is used in all three DGP's under the fuzzy design, the estimation results are averaged across these experiments.} The results are reported in Table A.2. The GP estimator  outperforms the LL estimators in terms of precision, coverage probability and interval length. 

\begin{table}[h!]
	%\caption{SRDP estimation results}\label{tab:classification}
	\centering
	\small
	Table A.2	SRDP estimation results on treatment take-up probability\\
	\vskip .1in
	\begin{tabular}{lcccccc}
		\hline
		&\multicolumn{3}{c}{$ N=300 $}&\multicolumn{3}{c}{$ N=500 $}\\\cline{2-7}
		&LL1&LL2&GP1&LL1&LL2&GP1\\\hline
		$ |\textnormal{Bias}| $&0.131&0.153&\textbf{0.056}&0.105&0.124&\textbf{0.047}\\
		RMSE&0.173&0.201&\textbf{0.070}&0.136&0.161&\textbf{0.059}\\
		Coverage&0.827&0.817&\textbf{0.947}&0.880&0.873&\textbf{0.960}\\
		IL&0.567&0.676&\textbf{0.298}&0.457&0.543&\textbf{0.259}\\
		\hline
	\end{tabular}
\end{table}

\subsection*{D. Covariate-adjusted estimation}	
We outline in this section the procedure to implement  covariate-adjusted GP regression under sharp RD design. To incorporate covariate $ z $, we generalize the `unadjusted' model as follows:
\begin{equation*}
	y_{i}=f(x_{i})+f_z(z_{i})+u_{i},i=1,\dots,N
\end{equation*}
To ease notation, we drop the subscript $ j=0,1 $ for the control and treatment groups in this section, with the understanding that this estimator is to be applied to both groups.
Following the suggestion of \cite{calonico2019regression}, we enter the covariate additively in a linear-in-parameter manner. This is achieved via a polynomial covariance function for covariate $ z $, yielding $ f_z(z)=\mb{h}^T_z(z)\beta_z $, where $ \mb{h}_z $ is a collection of polynomial basis functions and $ \beta_z $ is a vector of compatible dimension. Assuming a Gaussian prior $ \beta_z \sim \mathcal N(0,\Sigma_z) $, we have $ f_z(z) \sim \mathcal N(0,\mb{h}_z^T(z)\Sigma_z \mb h_z(z)) $. It follows that $ y|x,z \sim \mathcal N (0, k(x,x';\theta_x)+\mb h_z^T(z)\Sigma_z \mb h_z(z')+\sigma^2) $, where $ k $ is a generic covariance function with parameters $ \theta_x $. Let $ \mb H_z(\mb z) $ be the collection of $ \mb h_z $ and $ \mb K_{\mb x,\mb x}=k(\mb x,\mb x;\theta_x) $ given a sample $ \mathcal D=(\mb y, \mb x, \mb z) $. Define 
$ \mb K_f= \mb K_{\mb x,\mb x}+\sigma^2 \mathcal I_N$ and $ \bar \beta_z=\left\{\Sigma_z^{-1}+\mb H_z^T \mb K_f^{-1}\mb H_z \right\}^{-1}\left\{\mb H^T_z \mb K_f^{-1} \mb y\right\}  $. We can show that 
$	f_*|\mathcal D,x_*,\theta_x,\Sigma_z \sim \mathcal N(m_*,v_*)
$ with
\begin{align*}
	m_*&=\mb K_{x_*,\mb x}\mb K_f^{-1}(\mb y- \mb H_z(\mb z)\bar \beta_z)\\
	v_*&=K_{x_*,x_*}-\mb K_{x_*,\mb x}\mb K_f^{-1} \mb K_{\mb x,x_*}
\end{align*}
Comparison of these quantities with their `unadjusted' counterparts \eqref{fixed_predictive} suggests that the influence of covariate $ z $ has been profiled out in the predictive mean calculation. Although the formula for the predictive variance remains the same, note that $ \theta_x $, the parameters for the covariance function, are generally altered with the incorporation of covariate $ z $. Therefore, the predictive variance is also adjusted accordingly.

We apply this estimator to the control and treatment groups alike in our covariate-adjusted RD estimation. Tuning parameters for the running variable $ x $ and covariate $ z $ are learned jointly during the estimation process. Once the potential influence of the covariate is profiled out, the same produces for the `unadjusted' estimators are used for the estimation and inference of covariate-adjusted RD effect.
	
\end{appendix}

% \begin{thebibliography}{}
	% \bibitem{b1}
	% \end{thebibliography}

%\newpage
\bibliographystyle{asa} 
\bibliography{rd}

\end{document}